\documentclass[conference]{IEEEtran}
\IEEEoverridecommandlockouts
\pdfoutput=1
\usepackage{cite}
\usepackage{amsmath,amssymb,amsfonts}
\usepackage{algorithmic}
\usepackage{graphicx}
\usepackage{textcomp}
\usepackage{xcolor}
\def\BibTeX{{\rm B\kern-.05em{\sc i\kern-.025em b}\kern-.08em
    T\kern-.1667em\lower.7ex\hbox{E}\kern-.125emX}}
\begin{document}

\title{Creating Trustworthy LLMs: Dealing with Hallucinations in Healthcare AI}

\author{\IEEEauthorblockN{Muhammad Aurangzeb Ahmad}
\IEEEauthorblockA{\textit{Department of Computer Science} \\
\textit{University of Washington Bothell}\\
Bothell, WA\\
maahmad@uw.edu}
\and
\IEEEauthorblockN{Ilker Yaramis}
\IEEEauthorblockA{\textit{ACE AI (of KPInsight) } \\
\textit{Kaiser Permanente}\\
Oakland, USA \\
ilker.yaramis@kp.org}
\and
\IEEEauthorblockN{Taposh Dutta Roy}
\IEEEauthorblockA{\textit{ACE AI (of KPInsight)} \\
\textit{Kaiser Permanente}\\
Oakland, USA \\
taposh.d.roy@kp.org}
}

\maketitle

\begin{abstract}
Large language models have proliferated across multiple domains in as short period of time. There is however hesitation in the medical and healthcare domain towards their adoption because of issues like factuality, coherence, and hallucinations. Give the high stakes nature of healthcare, many researchers have even cautioned against its usage until these issues are resolved. The key to the implementation and deployment of LLMs in healthcare is to make these models trustworthy, transparent (as much possible)  and explainable. In this paper we describe the key elements in creating reliable, trustworthy, and unbiased models as a necessary condition for their adoption in healthcare. Specifically we focus on the quantification, validation, and mitigation of hallucinations in the context in healthcare. Lastly, we discuss how the future of LLMs in healthcare may look like.
\end{abstract}

\begin{IEEEkeywords}
LLM, AI Hallucination, ChatGPT
\end{IEEEkeywords}

\section{Introduction}
\noindent The proliferation of large language models (LLMs) is opening new challenges for their utilization in mission critical and high risk industries like healthcare, law, and behavioral therapy. LLMs have gone from models mainly confined to certain segments of industries to a technology which poised to penetrate almost every industry and domain. Rapid advances in LLMs like like BERT \cite{devlin2018bert}, BART \cite{lewis2019bart}, GPT-3 \cite{brown2020language}, GPT-4 \cite{sanderson2023gpt} and ChatGPT have dazzled the world oft times in generating impressive texts \cite{shen2023anything}, passing bar exams \cite{choi2023chatgpt}, outperforming humans in certain tasks \cite{lin2021truthfulqa}. One the flip-side, these models are also marred by issues like bias, privacy, security, and hallucinations. Like all impactful technologies, LLMs have unintended consequences \cite{li2023chatgpt} which are still being explored \cite{shen2023anything}. Hallucinating incorrect answers can have adverse monetary consequences e.g., when Google Bard hallucination at its launch cost the company \$100 Billion \cite{independentGPT23} and social consequences e.g., when ChatGPT falsely accused professor on sexually assaulting his students \cite{turley2023chatgpt}. This is one of the reasons why some researchers have recommended that LLMs should not be used in healthcare and medicine \cite{emsley2023chatgpt} \cite{salvagno2023artificial}. Another group of researchers recommend taking a more cautionary approach \cite{alkaissi2023artificial} towards adoption of LLMs. In this paper we argue that caution in warranted in the use of LLMs in healthcare, not using these models robs us for potential to address some pressing issues in healthcare AI. Instead, appropriate guardrails need to be in place before these models can live up to their true potential.

Creating trustworthy LLMs requires adhering to the principles of Responsible AI which requires ensuring that AI system are transparent, unbiased, accurate, interpretable, ensure privacy and security. In this paper we focus on AI hallucinations since they affect all of the elements of responsible AI and have been described as the most fundamental impediment in the widespread adoption on LLMs \cite{alkaissi2023artificial}. There are multiple competing and overlapping definitions of AI hallucinations depending upon the context. In general AI hallucinations refer to outputs from a LLM hat are contextually implausible \cite{lee2023benefits}, inconsistent with the real world and unfaithful to the input \cite{lee2023mathematical}. Some researchers have argued that the use of the term hallucination is a misnomer, it would be more accurate to describe AI Hallucinations as fabrications \cite{emsley2023chatgpt}.

In an ideal world LLM models would not produce any hallucinations. However, given how tokens are generated by LLMs, hallucinations are an inevitable end result of token generation in LLMs \cite{openai2023}. OpenAI, the creator of GPT-4, acknowledges on it website \cite{openai2023} hallucinations as a core limitation of LLMs. A recent study on LLMs for summarization demonstrated that hallucinated content was 25$\%$ of their generated summaries \cite{falke2019ranking}. In healthcare hallucinations can be specially problematic since the misinformation generated by LLMs could be related to diagnosis, treatment, or a recommended procedure. Training LLMs in the wild introduces additional issues like bias, non-factual data in addition to hallucinations. Thus, there are also calls for regulating the use of LLMs in healthcare \cite{mesko2023imperative}. AI Hallucinations happen in pretty much every modality where LLMs are applied e.g., text to text generation, text to image generation, text to video generation etc. While we discuss multiple modalities in this paper, we mainly focus on text related hallucinations in the application of AI.

\section{LLMs in Healthcare}
Popular LLMs like GPT-3, GPT-4, and ChatGPT have been used for multiple applications in healthcare. From an application perspective LLMs in healthcare have been explored for facilitating clinical workflow (e.g., recommendation, write discharge summary etc). translation, triage (guide patients to the right department, medical research (medical writing, anonymization etc), medical education (compose medical questions) Even though these models have been trained on general data, they perform relatively well on certain healthcare and medical tasks \cite{li2023chatgpt}. However for certain specialized tasks these models perform abysmally or generate incorrect information. Researchers have suggested \cite{singhal2023towards} that this limitation can be overcome by using datasets from healthcare domains e.g., LLMs trained using EHR data like GatorTron \cite{yang2022gatortron} and \cite{kraljevicforesight}, or LLMs trained using medical datasets like Med-PaLM 2  \cite{singhal2023towards} and Flan-PaLM \cite{chung2022scaling}. Hospital systems and healthcare organizations are still hesitant to employ LLMs in production because of high cost and liability associated with getting the questions wrong.

While the text generated by LLMs can be quite good often times, it also has the tendency to regurgitate unreliable information that is present on online resources. \cite{lin2021truthfulqa} observed that LLMs are likely to generate false yet widely circulated information. In retrospect this is not surprising given the training data that is used to train LLMs. However, this becomes especially problematic in the healthcare domain when LLMs give health related advice or information about the medical domain. LLMs have also been used to see if they can pass medical exams in various locales where these models have passed the exams with relatively good performance e.g., United States Medical Licensing Examinations (USMLE) \cite{kung2023performance}, Chinese National Medical Licensing Examination \cite{liu2023benchmarking}, and Japanese national medical licensing examinations \cite{kasai2023evaluating}. 

\begin{table*}[]
\centering
\begin{tabular}{ll}
\textbf{Area} & \textbf{Example} \\ \hline
Text Summarization & Generate summary of medical records or medical history \\
Annonymization & Annonymize patient data for privacy and HIPPA compliance \\
Clinical Documentation & Generate patient discharge reports \\
Translation & Translate patient records from one language to another. Also, translate relevant medical information to other languages for patients\\
Medical Writing & Facilitate medical research by assisting in medical writing and research \\
Triage & Guide patients to the right wards or departments \\
Consultation & Recommendations for patient self care \\ \hline
\end{tabular}
\caption{Applications of LLMs in Healthcare}
\end{table*}

\section{Hallucinations in Healthcare AI}
Hallucinations could be generated because of a variety of reasons. Here are some prominent sources of hallucinations in the context of healthcare.
\begin{itemize}
    \item \textbf{Unreliable Sources:} If the data comes from a general source then it is likely to perpetuate commonly held misconceptions \cite{manakul2023selfcheckgpt}
    \item  \textbf{Probabilistic Generation:} Given the probabilistic nature of text generation, recombination of completely reliable texts can still lead to generation of false statements \cite{salvagno2023artificial}
    \item Biased Training Data: Sources that may be biased may lead to generating hallucinations \cite{mbakwe2023chatgpt}
    \item \textbf{Insufficient Context:} Text generated by LLMs is based on prompts. Lack of context could lead to text generation with little or no correspondence to what the end user is looking for. \cite{salvagno2023artificial}
    \item \textbf{Self-Contradictions:} LLMs are not good at sequential reasoning. This may lead to self-contradictions \cite{manakul2023selfcheckgpt}.
\end{itemize}

\section{Addressing the problem of Hallucinations}
Model hallucination offer a number of impediments when it comes to the usage of these models in healthcare, as described in the previous section. These obstacles are however not insurmountable and can be overcome. To address these problems one first needs to measure, evaluate, and then mitigate hallucinations.

\subsection{Evaluating Hallucinations}
Model evaluation for LLMs can be divided into two main scenarios, whether one has access to the model itself or alternatively merely to the outputs of the model. In the first case one can check the likelihood of generation of the text against the distribution of the corpus that was used to train the model. This is the scenario where the organization itself trains and builds the model using its own data.  In the context of healthcare, one cannot always assume access to the model as organizations may be using off the shelf LLM models like ChatGPT, GPT-3 or GPT-4. For such scenarios self-check by the LLMs \cite{manakul2023selfcheckgpt}. The main idea is that when the response from the LLM is sampled multiple times for a given concept, the responses are likely to diverge or even contradict one another in case of hallucinations \cite{manakul2023selfcheckgpt}.

In some healthcare applications interpretability and transparency of the models would be crucial \cite{ahmad2018interpretable} for the use case i.e., how did the LLM come up with a particular output? The LLM could also be asked how it come up with a certain conclusion. This can help establish the correctness and interpretation of the output \cite{lim2023benchmarking}. Evaluation of many LLMs in healthcare and medicine have focused on gauging performance against benchmark multiple-choice-question-(MCQ). The problem with such evaluation is that does not mimic real world use cases. This is illustrated by Hickam's Dictum which is a counterargument to the use of Occam's razor which states that that any given time there are multiple possible diagnosis given a set of symptoms \cite{borden2013hickam}. Consequently, one can argue that LLM benchmarks are insufficient because in the real-world physicians rarely have one diagnosis for a condition.

\subsection{Measuring Hallucinations}
The text generated by large language models is open-ended in nature and thus traditional metrics of evaluating machine learning models cannot be readily applied. Method used for evaluating hallucinations can be divided into two main categories: Human evaluation and automated evaluation.

\subsubsection{Human Evaluation}
These evaluation involve standardized ways to evaluate outputs from LLMs by manual human annotation \cite{lee2022factuality}. \cite{lin2021truthfulqa} propose a benchmark where the answers from a Q\& A system are evaluated by human annotators. FActScores \cite{min2023factscore} evaluates text generated by LLMs via an evaluation method that breaks a generation into atomic facts which are in turn evaluated by human evaluators. Majority voting for evaluating healthcare related answers generated by LLMs has been employed for myopia care\cite{lim2023benchmarking}, maternity \cite{lim2023benchmarking}, diabetes \cite{hulman2023chatgpt}, cancer \cite{chen2023utility}, infant care \cite{lim2023benchmarking} etc.

\subsubsection{Automatic Evaluation}
A number of automatic evaluation methods use human annotations as inputs for evaluation e.g., \cite{lin2021truthfulqa} uses human evaluation to determine the truth claims of generated answers, \cite{zha2023alignscore} describe a scoring function that computes information alignment between two arbitrary texts. Another scheme is to use external models to evaluate LLMs, a strategy adopted by FactScore \cite{min2023factscore}.

\subsubsection{Evaluation Metrics}
Some researchers use traditional classification metrics like precision, recall, and F-Score for reporting model performance for scenarios where the output from the models is either true or false \cite{mundler2023self}. Different tasks have different metrics e.g., Perplexity and Cross-Entropy Loss for language models \cite{manakul2023selfcheckgpt}, ROGUE for summarization \cite{afzal2023challenges}, BLEU  or Meteor for machine translation \cite{afzal2023challenges}. While these metrics do not directly measure hallucinations, a low score can be an indicator that the output is likely to have hallucinations.

\subsection{Mitigating Hallucinations}
Since sources of hallucinations can creep in any part of the life cycle of LLMs and the definition of what constitutes a hallucination can also vary, mitigation of Hallucinations in healthcare is a multi-pronged process. Some of the important strategies for mitigating hallucination as it pertain to healthcare use cases are as follows:

\subsubsection{Human-In-The-Loop (HITL)}
Having humans with domain expertise in various parts of the model development process can greatly help in reducing hallucinations downstream. This includes human is data annotation, data classification, oversight committees for auditing model outputs, suggestions corrections whenever needed etc. Some researchers have also suggested \cite{lim2023benchmarking} real-time supervision of LLMs where a human in the loop oversees the outputs of the LLM. While this may be necessary for certain high stakes use cases in healthcare it is not scalable as it is not feasible for even group of humans to do real-time check of thousands of outputs. An alternative way to address is for the LLM to flag outputs for which it is not confident and defer to human judgment. The input from humans can in turn be source of additional training data which would improve the model further. Lastly, the end user feedback can be incorporated on the LLM response e.g., how ChatGPT has the thumbs up/thumbs down option on generated responses.

\subsubsection{Algorithmic Corrections}
Traditional machine learning techniques like using regularization, adding penalty to loss function, can also be employed. This would make the LLMs generalize better and make them less likely to hallucinate \cite{manakul2023selfcheckgpt}.

\subsubsection{Fine Tuning}
Fine tuning LLMs requires adapting the LLM t specific tasks or domains. It may involve adapting the LLM to specific tasks or domains. The drawback of fine-tuning is that it is computationally expensive an may be expensive from a monetary perspective. The promise of fine-tuning in healthcare is that a model trained on data specific to healthcare is less likely to make up health related information. However, in practice this is not always true as researchers have noted that fine-tuning does not guarantee improvement in performance as there are examples to the contrary which have been reported in literature \cite{wang2022preserving}.

\subsubsection{Improving Prompts}

In cases where the model is not confident about the output and is likely to be hallucination then the LLM could output “I do not know” given an input prompt \cite{manakul2023selfcheckgpt}.

\subsection{Adversarial Training}
Since LLMs can be exploited by adversarial attacks, one way to mitigate for such attacks is intentionally exposed to adversarial examples during training to ensure that these models are not compromised \cite{kumar2023certifying}.

\subsection{Input Validation}
Checking the inputs for validity can also mitigate against hallucinations. This could be done by checking the input against known standards, or running the input through a separate model designed to detect adversarial inputs \cite{kumar2023certifying}.

\subsection{Memory Augmentation}
An external knowledge source can be encoded into a key-value memory which can be integrated with the LLM \cite{wu2022efficient}. However, this technique has only been demonstrated for relatively small medals like T5. While researcher have suggested that it may be possible to enhance the performance of LLMs, it has not been explored because of large memory requirements \cite{jha2023dehallucinating}.

\subsection{Model Choice}
Ever since the public release of GPT-2 a large number of LLMs have been publicly released. The performance of these models varies with respect to different aspects of AI hallucinations as described above: \cite{mundler2023self} observed that ChatGPT and GPT-4 are much better at catching self-contradictions as compared to Vicuna-13B.

\subsection{Benchmark Audits}
The work on benchmarks described in the evaluation section assumes the veracity of benchmarks. In their detailed study of knowledge-grounded conversational benchmarks \cite{dziri2022origin} found that the benchmarks themselves include incorrect information which in turn leads to hallucinations. More disturbing is the fact that if the dataset contains even a small number of hallucinations then this may shift the data distribution in a manner that is likely to generate a greater number of hallucinations. What this alludes to is that benchmarks used for testing AI systems in healthcare should be verified by human domain experts to ensure veracity. Additionally, any model generated content should only be added to the training set only after it has been scrutinized by multiple domain experts.

\section{Conclusion}
Although LLMs are increasingly being used on healthcare the community as a whole is cautious about their use because of certain limitations, the foremost among these is model hallucination. LMs may generate plausible-sounding but incorrect or misleading information, making it crucial for healthcare professionals to critically evaluate and validate the outputs. LLMs excel at natural language processing tasks, enabling them to analyze vast amounts of medical literature, patient records, and clinical notes. LLMs can assist in medical diagnosis by analyzing symptoms, predicting potential diseases, suggesting treatment options, and aiding in personalized medicine.  

Even for benchmarks where these models have shown to excel, researchers argue \cite{mbakwe2023chatgpt} that framing of medical knowledge as narrow set of options or multiple choice questions creates a framing of false certainty and thus not a true representation of how medicine is practiced. Even though the process of evaluating hallucinations in LLMs is still being standardized, tools for detecting hallucinations are already being released like  Nvidia’s NeMo Guardrails \cite{shen2023anything}. In healthcare Human-in-the-loop systems for building and validation of LLMs for high stakes tasks may be a necessity for the foreseeable future given the high-risk nature of the healthcare domain. For low stakes tasks automation may be achieveable with appropriate guardrails in place. Lastly, widespread adoption of LLMs will also have to overcome possible regulatory issues in the near future.

\end{document}